\documentclass{article}

\usepackage{nips13submit_e,times}
\usepackage{amsmath,amsfonts,amssymb}
\usepackage{booktabs}
\usepackage{epsfig}
\usepackage{graphicx}
\usepackage{verbatim}
\usepackage{amsmath}
\usepackage{textcomp}
\usepackage{amssymb}
\usepackage{stackengine}
\usepackage{url}

\usepackage{amsmath,amsfonts,amssymb}
\usepackage{multirow}
\usepackage{chngpage}

\usepackage{marginnote}
\usepackage{paralist}
\usepackage[normalem]{ulem}
\usepackage{sidecap}
\usepackage{float}
\usepackage{caption3}
\DeclareCaptionOption{parskip}[]{}
\usepackage[small]{caption}
\usepackage{rotating}
\usepackage{subfigure}

\usepackage{hyperref}
\usepackage{xcolor}
\definecolor{dark-red}{rgb}{0.4,0.15,0.15}
\definecolor{dark-blue}{rgb}{0.15,0.15,0.4}
\definecolor{medium-blue}{rgb}{0,0,0.5}
\hypersetup{
    colorlinks, linkcolor={black},
    citecolor={dark-blue}, urlcolor={medium-blue}
}

\title{Going deeper with convolutions}

\author{
Christian Szegedy\\
Google Inc.\\
\And
Wei Liu \\
University of North Carolina, Chapel Hill\\ 
\And
Yangqing Jia\\
Google Inc.\\
\And
Pierre Sermanet \\
Google Inc.\\
\And
Scott Reed\\
University of Michigan\\
\And
Dragomir Anguelov\\
Google Inc.\\
\And
Dumitru Erhan\\
Google Inc.\\
\And
Vincent Vanhoucke\\
Google Inc.\\
\And
Andrew Rabinovich\\
Google Inc.\\
}

\nipsfinalcopy 

\begin{document}

\maketitle

\newtheorem{theorem}{Theorem}[section]
\newtheorem{lemma}[theorem]{Lemma}
\newtheorem{proposition}[theorem]{Proposition}
\newtheorem{corollary}[theorem]{Corollary}

\definecolor{LimeGreen}{rgb}{0.1,0.9,0.1}
\definecolor{Maroon}{rgb}{0.9,0.1,0.1}

\begin{abstract}
We propose a deep convolutional neural network architecture codenamed “Inception”, which was responsible for setting the new state of the art for classification and detection in the ImageNet Large-Scale Visual Recognition Challenge 2014 (ILSVRC’14). The main hallmark of this architecture is the improved utilization of the computing resources inside the network. This was achieved by a carefully crafted design that allows for increasing the depth and width of the network while keeping the computational budget constant. To optimize quality, the architectural decisions were based on the Hebbian principle and the intuition of multi-scale processing. One particular incarnation used in our submission for ILSVRC’14 is called GoogLeNet, a 22 layers deep network, the quality of which is assessed in the context of classification and detection.
\end{abstract}

\section{Introduction}

In the last three years, mainly due to the advances of deep learning, more concretely convolutional networks ~\cite{lecun1989backprop}, the quality of image recognition and object detection has been progressing at a dramatic pace. One encouraging news is that most of this progress is not just the result of more powerful hardware, larger datasets and bigger models, but mainly a consequence of new ideas, algorithms and improved network architectures. No new data sources were used, for example, by the top entries in the ILSVRC 2014 competition besides the classification dataset of the same competition for detection purposes. Our GoogLeNet submission to ILSVRC 2014 actually uses $12\times$ fewer parameters than the winning architecture of Krizhevsky et al~\cite{krizhevsky2012imagenet} from two years ago, while being significantly more accurate. The biggest gains in object-detection have not come from the utilization of deep networks alone or bigger models, but from the synergy of deep architectures and classical computer vision, like the R-CNN algorithm by Girshick et al~\cite{girshick2014rich}. 

Another notable factor is that with the ongoing traction of mobile and embedded computing, the efficiency of our algorithms -- especially their power and memory use -- gains importance. It is noteworthy that the considerations leading to the design of the deep architecture presented in this paper included this factor rather than having a sheer fixation on accuracy numbers. For most of the experiments, the models were designed to keep a computational budget of $1.5$ billion multiply-adds at inference time, so that the they do not end up to be a purely academic curiosity, but could be put to real world use, even on large datasets, at a reasonable cost.

In this paper, we will focus on an efficient deep neural network architecture for computer vision, codenamed “Inception”, which derives its name from the “Network in network” paper by Lin et al~\cite{lin2013nin} in conjunction with the famous ``we need to go deeper'' internet meme~\cite{knowyourmeme}. In our case, the word ``deep'' is used in two different meanings: first of all, in the sense that we introduce a new level of organization in the form of the ``Inception module'' and also in the more direct sense of increased network depth. In general, one can view the Inception model as a logical culmination of ~\cite{lin2013nin} while taking inspiration and guidance from the theoretical work by Arora et al~\cite{arora2013bounds}. The benefits of the architecture are experimentally verified on the ILSVRC 2014 classification and detection challenges, on which it significantly outperforms the current state of the art. 

\section{Related Work}

Starting with LeNet-5 ~\cite{lecun1989backprop}, convolutional neural networks (CNN) have typically had a standard structure -- stacked convolutional layers (optionally followed by contrast normalization and max-pooling) are followed by one or more fully-connected layers. Variants of this basic design are prevalent in the image classification literature and have yielded the best results to-date on MNIST, CIFAR and most notably on the ImageNet classification challenge~\cite{krizhevsky2012imagenet,zeiler2014visualizing}.  For larger datasets such as Imagenet, the recent trend has been to increase the number of layers ~\cite{lin2013nin} and layer size~\cite{zeiler2014visualizing,sermanet2013overfeat}, while using dropout~\cite{hinton2012dropout} to address the problem of overfitting.

Despite concerns that max-pooling layers result in loss of accurate spatial information, the same convolutional network architecture as ~\cite{krizhevsky2012imagenet} has also been successfully employed for localization~\cite{krizhevsky2012imagenet,sermanet2013overfeat}, object detection~\cite{girshick2014rich,sermanet2013overfeat,szegedy2013deep,erhan2014scalable} and human pose estimation~\cite{toshev2013deep}.
Inspired by a neuroscience model of the primate visual cortex, Serre et al.~\cite{serre2007robust} use a series of fixed Gabor filters of different sizes in order to handle multiple scales, similarly to the Inception model. However, contrary to the fixed 2-layer deep model of ~\cite{serre2007robust}, all filters in the Inception model are learned. Furthermore, Inception layers are repeated many times, leading to a 22-layer deep model in the case of the GoogLeNet model.

Network-in-Network is an approach proposed by Lin et al.~\cite{lin2013nin} in order to increase the representational power of neural networks. When applied to convolutional layers, the method could be viewed as additional $1\times 1$ convolutional layers followed typically by the rectified linear activation ~\cite{krizhevsky2012imagenet}. This enables it to be easily integrated in the current CNN pipelines. We use this approach heavily in our architecture. However, in our setting, $1\times 1$ convolutions have dual purpose: most critically, they are used mainly as dimension reduction modules to remove computational bottlenecks, that would otherwise limit the size of our networks.  This allows for not just increasing the depth, but also the width of our networks without significant performance penalty.

The current leading approach for object detection is the Regions with Convolutional Neural Networks (R-CNN) proposed by Girshick et al.~\cite{girshick2014rich}. R-CNN decomposes the overall detection problem into two subproblems: to first utilize low-level cues such as color and superpixel consistency for potential object proposals in a category-agnostic fashion, and to then use CNN classifiers to identify object categories at those locations. Such a two stage approach leverages the accuracy of bounding box segmentation with low-level cues, as well as the highly powerful classification power of state-of-the-art CNNs. We adopted a similar pipeline in our detection submissions, but have explored enhancements in both stages, such as multi-box ~\cite{erhan2014scalable} prediction for higher object bounding box recall, and ensemble approaches for better categorization of bounding box proposals.

\section{Motivation and High Level Considerations}

The most straightforward way of improving the performance of deep neural networks is by increasing their size. This includes both increasing the depth -- the number of levels -- of the network and its width: the number of units at each level. This is as an easy and safe way of training higher quality models, especially given the availability of a large amount of labeled training data. However this simple solution comes with two major drawbacks. 

\begin{figure}
\centering
\subfigure[Siberian husky]{
\includegraphics[width=0.49\textwidth]{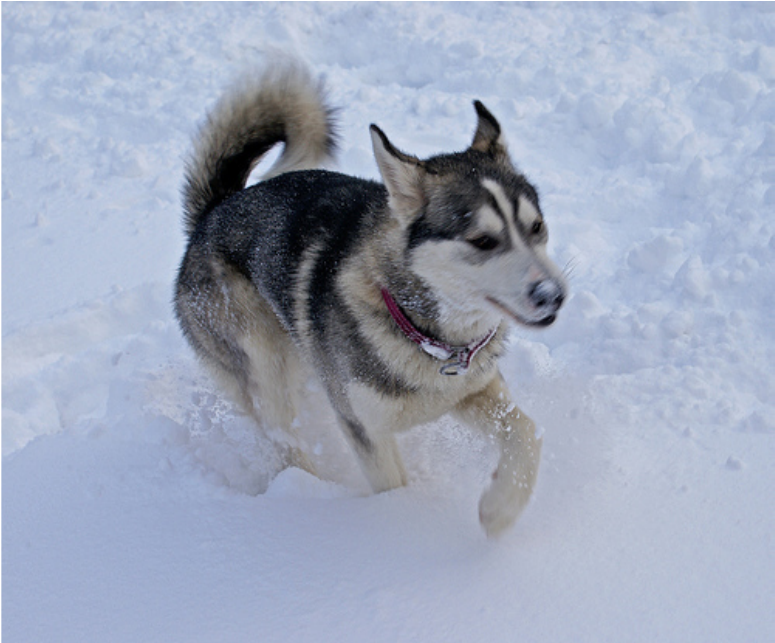}
\label{fig:siberianhusky}
}
\subfigure[Eskimo dog]{
\includegraphics[width=0.4658\textwidth]{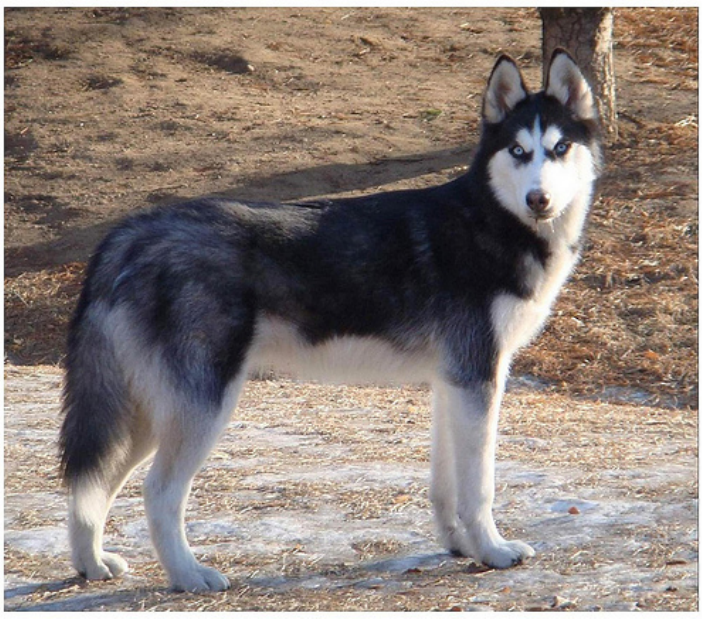}
\label{fig:eskimodog}
}
\caption{Two distinct classes from the 1000 classes of the ILSVRC 2014 classification challenge.}
\label{fig:finegrained}
\end{figure}

Bigger size typically means a larger number of parameters, which makes the enlarged network more prone to overfitting, especially if the number of labeled examples in the training set is limited. This can become a major bottleneck, since the creation of high quality training sets can be tricky and expensive, especially if expert human raters are necessary to distinguish between fine-grained visual categories like those in ImageNet (even in the 1000-class ILSVRC subset) as demonstrated by Figure~\ref{fig:finegrained}.

Another drawback of uniformly increased network size is the dramatically increased use of computational resources. For example, in a deep vision network, if two convolutional layers are chained, any uniform increase in the number of their filters results in a quadratic increase of computation. If the added capacity is used inefficiently (for example, if most weights end up to be close to zero), then a lot of computation is wasted. Since in practice the computational budget is always finite, an efficient distribution of computing resources is preferred to an indiscriminate increase of size, even when the main objective is to increase the quality of results.

The fundamental way of solving both issues would be by ultimately moving from fully connected to sparsely connected architectures, even inside the convolutions. Besides mimicking biological systems, this would also have the advantage of firmer theoretical underpinnings due to the groundbreaking work of Arora et al.~\cite{arora2013bounds}. Their main result states that if the probability distribution of the data-set is representable by a large, very sparse deep neural network, then the optimal network topology can be constructed layer by layer by analyzing the correlation statistics of the activations of the last layer and clustering neurons with highly correlated outputs. Although the strict mathematical proof requires very strong conditions, the fact that this statement resonates with the well known Hebbian principle -- neurons that fire together, wire together -- suggests that the underlying idea is applicable even under less strict conditions, in practice.

On the downside, today’s computing infrastructures are very inefficient when it comes to numerical calculation on non-uniform sparse data structures. Even if the number of arithmetic operations is reduced by $100{\times}$, the overhead of lookups and cache misses is so dominant that switching to sparse matrices would not pay off. The gap is widened even further by the use of steadily improving, highly tuned, numerical libraries that allow for extremely fast dense matrix multiplication, exploiting the minute details of the underlying CPU or GPU hardware~\cite{song2014scaling, krizhevsky2012imagenet}. Also, non-uniform sparse models require more sophisticated engineering and computing infrastructure. Most current vision oriented machine learning systems utilize sparsity in the spatial domain just by the virtue of employing convolutions. However, convolutions are implemented as collections of dense connections to the patches in the earlier layer. ConvNets have traditionally used random and sparse connection tables in the feature dimensions since~\cite{lecun1998gradient} in order to break the symmetry and improve learning, the trend changed back to full connections with \cite{krizhevsky2012imagenet} in order to better optimize parallel computing. The uniformity of the structure and a large number of filters and greater batch size allow for utilizing efficient dense computation.

This raises the question whether there is any hope for a next, intermediate step: an architecture that makes use of the extra sparsity, even at filter level, as suggested by the theory, but exploits our current hardware by utilizing computations on dense matrices. The vast literature on sparse matrix computations (e.g.~\cite{catalyurek2010sparse}) suggests that clustering sparse matrices into relatively dense submatrices tends to give state of the art practical performance for sparse matrix multiplication. It does not seem far-fetched to think that similar methods would be utilized for the automated construction of non-uniform deep-learning architectures in the near future.

The Inception architecture started out as a case study of the first author for assessing the hypothetical output of a sophisticated network topology construction algorithm that tries to approximate a sparse structure implied by \cite{arora2013bounds} for vision networks and covering the hypothesized outcome by dense, readily available components. Despite being a highly speculative undertaking, only after two iterations on the exact choice of topology, we could already see modest gains against the reference architecture based on \cite{lin2013nin}. After further tuning of learning rate, hyperparameters and improved training methodology, we established that the resulting Inception architecture was especially useful in the context of localization and object detection as the base network for~\cite{girshick2014rich} and ~\cite{erhan2014scalable}. Interestingly, while most of the original architectural choices have been questioned and tested thoroughly, they turned out to be at least locally optimal.

One must be cautious though: although the proposed architecture has become a success for computer vision, it is still questionable whether its quality can be attributed to the guiding principles that have lead to its construction. Making sure would require much more thorough analysis and verification: for example, if automated tools based on the principles described below would find similar, but better topology for the vision networks. The most convincing proof would be if an automated system would create network topologies resulting in similar gains in other domains using the same algorithm but with very differently looking global architecture. At very least, the initial success of the Inception architecture yields firm motivation for exciting future work in this direction.

\section{Architectural Details}

\begin{figure}
\centering
\subfigure[Inception module, na\"{i}ve version]{
\includegraphics[width=0.48\textwidth]{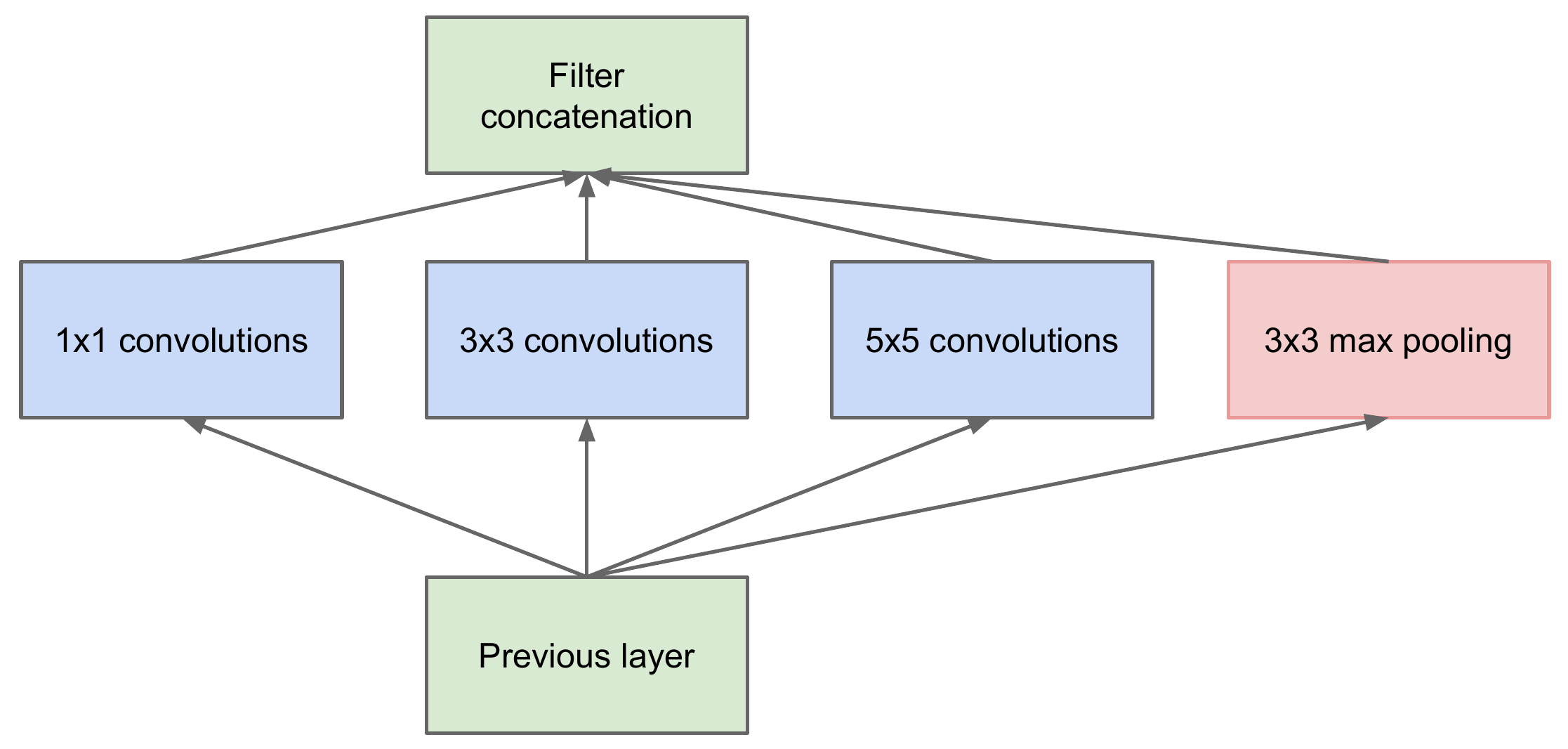}
\label{fig:inceptionnaive}
}
\subfigure[Inception module with dimension reductions]{
\includegraphics[width=0.48\textwidth]{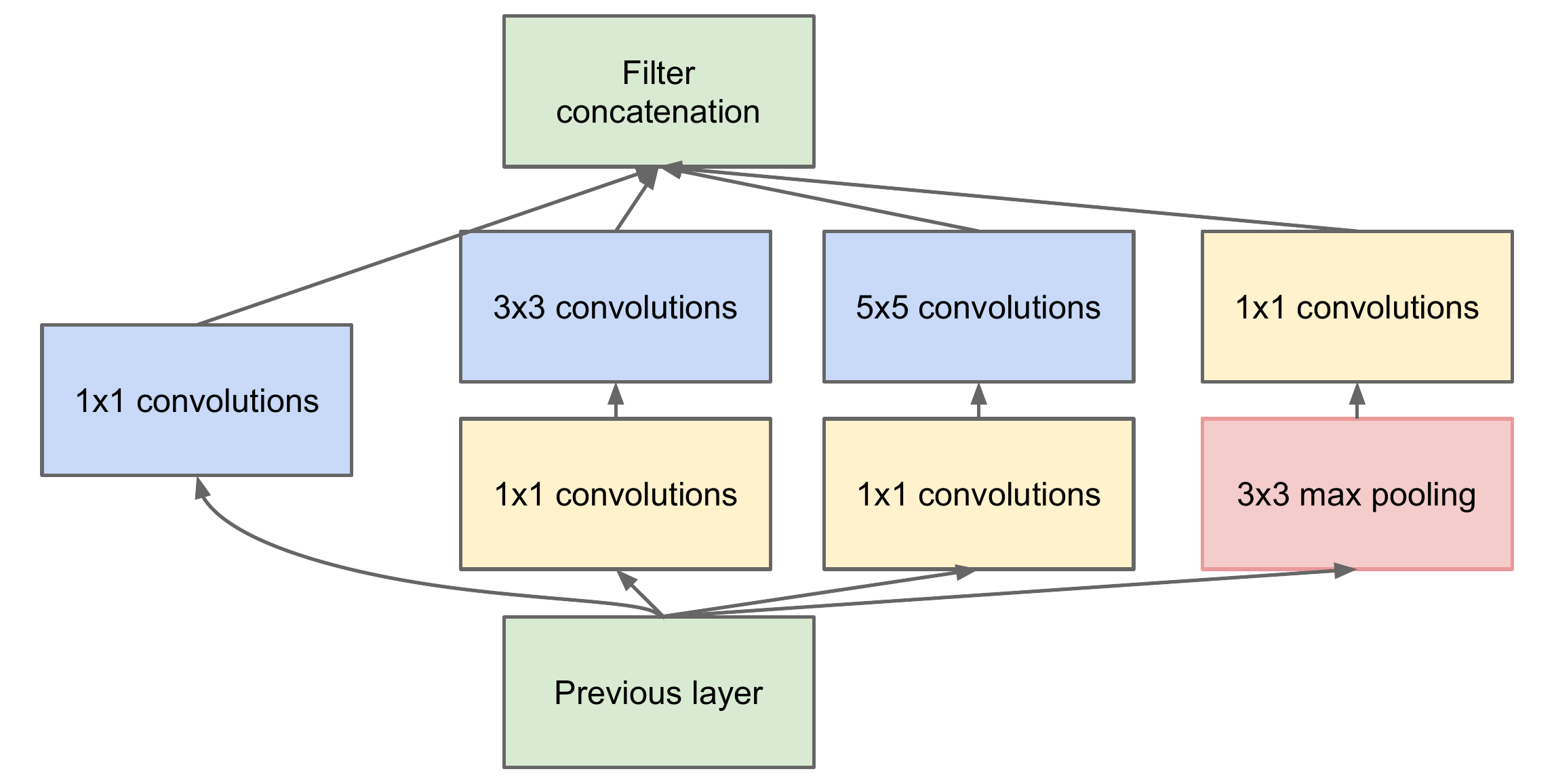}
\label{fig:inceptionmodule}
}
\caption{Inception module}
\label{fig:inception}
\end{figure}

The main idea of the Inception architecture is based on finding out how an optimal local sparse structure in a convolutional vision network can be approximated and covered by readily available dense components. Note that assuming translation invariance means that our network will be built from convolutional building blocks. All we need is to find the optimal local construction and to repeat it spatially. Arora et al.~\cite{arora2013bounds} suggests a layer-by layer construction in which one should analyze the correlation statistics of the last layer and cluster them into groups of units with high correlation. These clusters form the units of the next layer and are connected to the units in the previous layer. We assume that each unit from the earlier layer corresponds to some region of the input image and these units are grouped into filter banks. In the lower layers (the ones close to the input) correlated units would concentrate in local regions. This means, we would end up with a lot of clusters concentrated in a single region and they can be covered by a layer of $1{\times}1$ convolutions in the next layer, as suggested in ~\cite{lin2013nin}. However, one can also expect that there will be a smaller number of more spatially spread out clusters that can be covered by convolutions over larger patches, and there will be a decreasing number of patches over larger and larger regions. In order to avoid patch-alignment issues, current incarnations of the Inception architecture are restricted to filter sizes $1{\times}1$, $3{\times}3$ and $5{\times}5$, however this decision was based more on convenience rather than necessity. It also means that the suggested architecture is a combination of all those layers with their output filter banks concatenated into a single output vector forming the input of the next stage. Additionally, since pooling operations have been essential for the success in current state of the art convolutional networks, it suggests that adding an alternative parallel pooling path in each such stage should have additional beneficial effect, too (see Figure~\ref{fig:inceptionnaive}).

As these ``Inception modules'' are stacked on top of each other, their output correlation statistics are bound to vary: as features of higher abstraction are captured by higher layers, their spatial concentration is expected to decrease suggesting that the ratio of $3{\times}3$ and $5{\times}5$ convolutions should increase as we move to higher layers. 

One big problem with the above modules, at least in this na\"{i}ve form, is that even a modest number of $5{\times}5$ convolutions can be prohibitively expensive on top of a convolutional layer with a large number of filters. This problem becomes even more pronounced once pooling units are added to the mix: their number of output filters equals to the number of filters in the previous stage. The merging of the output of the pooling layer with the outputs of convolutional layers would lead to an inevitable increase in the number of outputs from stage to stage. Even while this architecture might cover the optimal sparse structure, it would do it very inefficiently, leading to a computational blow up within a few stages. 

This leads to the second idea of the proposed architecture: judiciously applying dimension reductions and projections wherever the computational requirements would increase too much otherwise. This is based on the success of embeddings: even low dimensional embeddings might contain a lot of information about a relatively large image patch. However, embeddings represent information in a dense, compressed form and compressed information is harder to model. We would like to keep our representation sparse at most places (as required by the conditions of ~\cite{arora2013bounds}) and compress the signals only whenever they have to be aggregated en masse. That is, $1{\times}1$ convolutions are used to compute reductions before the expensive $3{\times}3$ and $5{\times}5$ convolutions. Besides being used as reductions, they also include the use of rectified linear activation which makes them dual-purpose. The final result is depicted in Figure~\ref{fig:inceptionmodule}.

In general, an Inception network is a network consisting of modules of the above type stacked upon each other, with occasional max-pooling layers with stride 2 to halve the resolution of the grid. For technical reasons (memory efficiency during training), it seemed beneficial to start using Inception modules only at higher layers while keeping the lower layers in traditional convolutional fashion. This is not strictly necessary, simply reflecting some infrastructural inefficiencies in our current implementation.

One of the main beneficial aspects of this architecture is that it allows for increasing the number of units at each stage significantly without an uncontrolled blow-up in computational complexity. The ubiquitous use of dimension reduction allows for shielding the large number of input filters of the last stage to the next layer, first reducing their dimension before convolving over them with a large patch size. Another practically useful aspect of this design is that it aligns with the intuition that visual information should be processed at various scales and then aggregated so that the next stage can abstract features from different scales simultaneously.

The improved use of computational resources allows for increasing both the width of each stage as well as the number of stages without getting into computational difficulties. Another way to utilize the inception architecture is to create slightly inferior, but computationally cheaper versions of it. We have found that all the included the knobs and levers allow for a controlled balancing of computational resources that can result in networks that are $2-3{\times}$ faster than similarly performing networks with non-Inception architecture, however this requires careful manual design at this point.

\section{GoogLeNet}

We chose GoogLeNet as our team-name in the ILSVRC14 competition. This name is an homage to Yann LeCun’s pioneering LeNet 5 network~\cite{lecun1989backprop}. We also use GoogLeNet to refer to the particular incarnation of the Inception architecture used in our submission for the competition. We have also used a deeper and wider Inception network, the quality of which was slightly inferior, but adding it to the ensemble seemed to improve the results marginally. We omit the details of that network, since our experiments have shown that the influence of the exact architectural parameters is relatively minor. Here, the most successful particular instance (named GoogLeNet) is described in Table \ref{googlenet} for demonstrational purposes. The exact same topology (trained with different sampling methods) was used for 6 out of the 7 models in our ensemble.

\begin{table}
{\tiny
\begin{tabular}[H]{|l|c|c|c|c|c|c|c|c|c|c|c|}
\hline
{\bf type} & {\bf \stackanchor{patch size/}{stride}} & {\bf \stackanchor{output}{size}} & 
{\bf depth} & {\bf $\#1{\times}1$} & {\bf \stackanchor{$\#3{\times}3$}{reduce}} & $\#3{\times}3$ & 
{\bf \stackanchor{$\#5{\times}5$}{reduce}} & $\#5{\times}5$ & {\bf \stackanchor{pool}{proj}} & 
{\bf params} & {\bf ops} \\
\hline\hline
convolution & $7{\times}7/2$ & $112{\times}112{\times}64$ & 1 & & & & & & & 2.7K & 34M \\
\hline
max pool & $3{\times}3/2$ & $56{\times}56{\times}64$ & 0 & & & & & & & & \\
\hline
convolution & $3{\times}3/1$ & $56{\times}56{\times}192$ & 2 & & 64 & 192 & & & & 112K & 360M \\
\hline
max pool & $3{\times}3/2$ & $28{\times}28{\times}192$ & 0 & & & & & & & & \\
\hline
inception (3a) & & $28{\times}28{\times}256$ & 2 & 64 & 96 & 128 & 16 & 32 & 32 & 159K & 128M \\
\hline
inception (3b) & & $28{\times}28{\times}480$ & 2 & 128 & 128 & 192 & 32 & 96 & 64 & 380K & 304M \\
\hline
max pool & $3{\times}3/2$ & $14{\times}14{\times}480$ & 0 & & & & & & & & \\
\hline
inception (4a) & & $14{\times}14{\times}512$ & 2 & 192 & 96 & 208 & 16 & 48 & 64 & 364K & 73M \\
\hline
inception (4b) & & $14{\times}14{\times}512$ & 2 & 160 & 112 & 224 & 24 & 64 & 64 & 437K & 88M \\
\hline
inception (4c) & & $14{\times}14{\times}512$ & 2 & 128 & 128 & 256 & 24 & 64 & 64 & 463K & 100M \\
\hline
inception (4d) & & $14{\times}14{\times}528$ & 2 & 112 & 144 & 288 & 32 & 64 & 64 & 580K & 119M \\
\hline
inception (4e) & & $14{\times}14{\times}832$ & 2 & 256 & 160 & 320 & 32 & 128 & 128 & 840K & 170M \\
\hline
max pool & $3{\times}3/2$ & $7{\times}7{\times}832$ & 0 & & & & & & & & \\
\hline
inception (5a) & & $7{\times}7{\times}832$ & 2 & 256 & 160 & 320 & 32 & 128 & 128 & 1072K & 54M \\
\hline
inception (5b) & & $7{\times}7{\times}1024$ & 2 & 384 & 192 & 384 & 48 & 128 & 128 & 1388K & 71M \\
\hline
avg pool & $7{\times}7/1$ & $1{\times}1{\times}1024$ & 0 & & & & & & & & \\
\hline
dropout (40\%) & & $1{\times}1{\times}1024$ & 0 & & & & & & & & \\
\hline
linear & & $1{\times}1{\times}1000$ & 1 & & & & & & & 1000K & 1M \\
\hline
softmax & & $1{\times}1{\times}1000$ & 0 & & & & & & & & \\
\hline
\end{tabular}
}
\caption{GoogLeNet incarnation of the Inception architecture}
\label{googlenet}
\end{table}

All the convolutions, including those inside the Inception modules, use rectified linear activation. The size of the receptive field in our network is $224{\times}224$ taking RGB color channels with mean subtraction. ``$\#3{\times}3$ reduce'' and ``$\#5{\times}5$ reduce'' stands for the number of $1{\times}1$ filters in the reduction layer used before the $3{\times}3$ and $5{\times}5$ convolutions. One can see the number of $1{\times}1$ filters in the projection layer after the built-in max-pooling in the “pool proj” column. All these reduction/projection layers use rectified linear activation as well.

\begin{figure}
\centering
\includegraphics[width=0.38\textwidth]{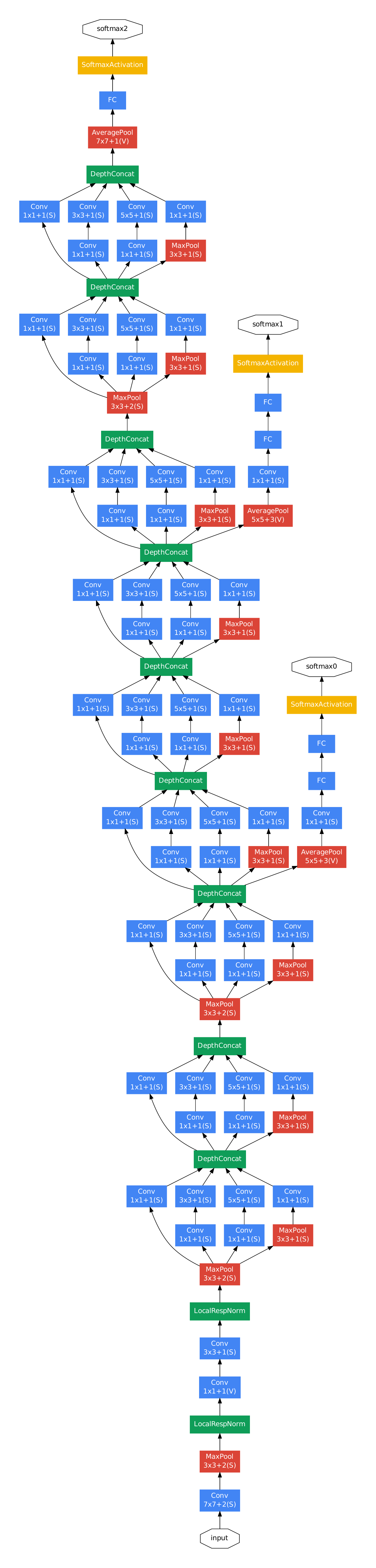}
\caption{GoogLeNet network with all the bells and whistles}
\label{fig:googlenet}
\end{figure}

The network was designed with computational efficiency and practicality in mind, so that inference can be run on individual devices including even those with limited computational resources, especially with low-memory footprint. The network is 22 layers deep when counting only layers with parameters (or 27 layers if we also count pooling). The overall number of “layers” (independent building blocks) used for the construction of the network is about 100. However this number depends on the machine learning infrastructure system used. The use of average pooling before the classifier is based on \cite{lin2013nin}, although our implementation differs in that we use an extra linear layer. This enables adapting and fine-tuning our networks for other label sets easily, but it is mostly convenience and we do not expect it to have a major effect. It was found that a move from fully connected layers to average pooling improved the top-1 accuracy by about 0.6\%, however the use of dropout remained essential even after removing the fully connected layers.

Given the relatively large depth of the network, the ability to propagate gradients back through all the layers in an effective manner was a concern. One interesting insight is that the strong performance of relatively shallower networks on this task suggests that the features produced by the layers in the middle of the network should be very discriminative. By adding auxiliary classifiers connected to these intermediate layers, we would expect to encourage discrimination in the lower stages in the classifier, increase the gradient signal that gets propagated back, and provide additional regularization. These classifiers take the form of smaller convolutional networks put on top of the output of the Inception (4a) and (4d) modules. During training, their loss gets added to the total loss of the network with a discount weight (the losses of the auxiliary classifiers were weighted by 0.3). At inference time, these auxiliary networks are discarded.

The exact structure of the extra network on the side, including the auxiliary classifier, is as follows:
\begin{itemize}
\item An average pooling layer with $5{\times}5$ filter size and stride $3$, resulting in an $4{\times}4{\times}512$ output for the (4a), and $4{\times}4{\times}528$ for the (4d) stage.
\item A $1{\times}1$ convolution with 128 filters for dimension reduction and rectified linear activation.
\item A fully connected layer with 1024 units and rectified linear activation.
\item A dropout layer with 70\% ratio of dropped outputs.
\item A linear layer with softmax loss as the classifier (predicting the same 1000 classes as the main classifier, but removed at inference time). 
\end{itemize}

A schematic view of the resulting network is depicted in Figure~\ref{fig:googlenet}.

\section{Training Methodology}

Our networks were trained using the DistBelief~\cite{dean2012large} distributed machine learning system using modest amount of model and data-parallelism. Although we used CPU based implementation only, a rough estimate suggests that the GoogLeNet network could be trained to convergence using few high-end GPUs within a week, the main limitation being the memory usage. Our training used asynchronous stochastic gradient descent with 0.9 momentum~\cite{sutskever2013momentum}, fixed learning rate schedule (decreasing the learning rate by 4\% every 8 epochs). Polyak averaging~\cite{polyak1992} was used to create the final model used at inference time.

  Our image sampling methods have changed substantially over the months leading to the competition, and already converged models were trained on with other options, sometimes in conjunction with changed hyperparameters, like dropout and learning rate, so it is hard to give a definitive guidance to the most effective single way to train these networks. To complicate matters further, some of the models were mainly trained on smaller relative crops, others on larger ones, inspired by~\cite{howard2013improvements}. Still, one prescription that was verified to work very well after the competition includes sampling of various sized patches of the image whose size is distributed evenly between 8\% and 100\% of the image area and whose aspect ratio is chosen randomly between $3/4$ and $4/3$. Also, we found that the photometric distortions by Andrew Howard~\cite{howard2013improvements} were useful to combat overfitting to some extent. In addition, we started to use random interpolation methods (bilinear, area, nearest neighbor and cubic, with equal probability) for resizing relatively late and in conjunction with other hyperparameter changes, so we could not tell definitely whether the final results were affected positively by their use.

\section{ILSVRC 2014 Classification Challenge Setup and Results}

The ILSVRC 2014 classification challenge involves the task of classifying the image into one of 1000 leaf-node categories in the Imagenet hierarchy. There are about 1.2 million images for training, 50,000 for validation and 100,000 images for testing. Each image is associated with one ground truth category, and performance is measured based on the highest scoring classifier predictions. Two numbers are usually reported: the top-1 accuracy rate, which compares the ground truth against the first predicted class, and the top-5 error rate, which compares the ground truth against the first 5 predicted classes: an image is deemed correctly classified if the ground truth is among the top-5, regardless of its rank in them. The challenge uses the top-5 error rate for ranking purposes.

We participated in the challenge with no external data used for training. In addition to the training techniques aforementioned in this paper, we adopted a set of techniques during testing to obtain a higher performance, which we elaborate below.

\begin{enumerate}
\item We independently trained 7 versions of the same GoogLeNet model (including one wider version), and performed ensemble prediction with them. These models were trained with the same initialization (even with the same initial weights, mainly because of an oversight) and learning rate policies, and they only differ in sampling methodologies and the random order in which they see input images.
\item During testing, we adopted a more aggressive cropping approach than that of Krizhevsky et al.~\cite{krizhevsky2012imagenet}. Specifically, we resize the image to 4 scales where the shorter dimension (height or width) is 256, 288, 320 and 352 respectively, take the left, center and right square of these resized images (in the case of portrait images, we take the top, center and bottom squares). For each square, we then take the 4 corners and the center $224{\times}224$ crop as well as the square resized to $224{\times}224$, and their mirrored versions. This results in $4{\times}3{\times}6{\times}2=144$ crops per image. A similar approach was used by Andrew Howard~\cite{howard2013improvements} in the previous year's entry, which we empirically verified to perform slightly worse than the proposed scheme. We note that such aggressive cropping may not be necessary in real applications, as the benefit of more crops becomes marginal after a reasonable number of crops are present (as we will show later on).
\item The softmax probabilities are averaged over multiple crops and over all the individual classifiers to obtain the final prediction. In our experiments we analyzed alternative approaches on the validation data, such as max pooling over crops and averaging over classifiers, but they lead to inferior performance than the simple averaging.
\end{enumerate}

In the remainder of this paper, we analyze the multiple factors that contribute to the overall performance of the final submission.

\begin{table}
\centering
\begin{tabular}[H]{|l||l|l|l|l|}
\hline
{\bf Team} & {\bf Year} & {\bf Place} & {\bf Error (top-5)} & {\bf Uses external data} \\
\hline\hline
SuperVision & 2012 & 1st & $16.4\%$ & no \\
\hline
SuperVision & 2012 & 1st & $15.3\%$ & Imagenet $22$k \\
\hline
Clarifai & 2013 & 1st & $11.7\%$ & no \\
\hline
Clarifai & 2013 & 1st & $11.2\%$ &  Imagenet $22$k \\
\hline
MSRA & 2014 & 3rd & $7.35\%$ & no \\
\hline
VGG & 2014 & 2nd & $7.32\%$ & no \\
\hline
GoogLeNet & 2014 & 1st & $6.67\%$ & no\\
\hline
\end{tabular}
\caption{Classification performance}
\label{classfinal}
\end{table}
Our final submission in the challenge obtains a top-5 error of 6.67\% on both the validation and testing data, ranking the first among other participants. This is a 56.5\% relative reduction compared to the SuperVision approach in 2012, and about 40\% relative reduction compared to the previous year's best approach (Clarifai), both of which used external data for training the classifiers. The following table shows the statistics of some of the top-performing approaches.

\begin{table}
\centering
\begin{tabular}[H]{|l||l|l|l|l|}
\hline
{\bf Number of models} & {\bf Number of Crops} & {\bf Cost} & {\bf Top-5 error} & {\bf compared to base}\\
\hline
1 & 1 & 1 & $10.07\%$ & base \\
\hline
1 & 10 & 10 & $9.15\%$ & -0.92\%\\
\hline
1 & 144 & 144 & $7.89\%$ & -2.18\% \\
\hline
7 & 1 & 7 & $8.09\%$ & -1.98\% \\
\hline
7 & 10 & 70 & $7.62\%$ & -2.45\% \\
\hline
7 & 144 & 1008 & $6.67\%$ & -3.45\% \\
\hline
\end{tabular}
\caption{GoogLeNet classification performance break down}
\label{classbreakdown}
\end{table}

We also analyze and report the performance of multiple testing choices, by varying the number of models and the number of crops used when predicting an image in the following table. When we use one model, we chose the one with the lowest top-1 error rate on the validation data. All numbers are reported on the validation dataset in order to not overfit to the testing data statistics.

\section{ILSVRC 2014 Detection Challenge Setup and Results}

The ILSVRC detection task is to produce bounding boxes around objects in images among 200 possible classes. Detected objects count as correct if they match the class of the groundtruth and their bounding boxes overlap by at least 50\% (using the Jaccard index). Extraneous detections count as false positives and are penalized. Contrary to the classification task, each image may contain many objects or none, and their scale may vary from large to tiny. Results are reported using the mean average precision (mAP).

The approach taken by GoogLeNet for detection is similar to the R-CNN by \cite{girshick2014rich}, but is augmented with the Inception model as the region classifier. Additionally, the region proposal step is improved by combining the Selective Search~\cite{sande2011search} approach with multi-box~\cite{erhan2014scalable} predictions for higher object bounding box recall. In order to cut down the number of false positives, the superpixel size was increased by $2\times$. This halves the proposals coming from the selective search algorithm. We added back 200 region proposals coming from multi-box~\cite{erhan2014scalable} resulting, in total, in about 60\% of the proposals used by \cite{girshick2014rich}, while increasing the coverage from 92\% to 93\%. The overall effect of cutting the number of proposals with increased coverage is a 1\% improvement of the mean average precision for the single model case. Finally, we use an ensemble of 6 ConvNets when classifying each region which improves results from ~40\% to 43.9\% accuracy. Note that contrary to R-CNN, we did not use bounding box regression due to lack of time.

\begin{table}
\centering
\begin{tabular}[H]{|l||l|l|l|l|c|c|}
\hline
{\bf Team} & {\bf Year} & {\bf Place} & {\bf mAP} & {\bf external data} & {\bf ensemble} & {\bf approach} \\
\hline
UvA-Euvision & 2013 & 1st & $22.6\%$ & none & ? & Fisher vectors \\
\hline
Deep Insight & 2014 & 3rd & $40.5\%$ & ImageNet $1$k & 3 & CNN \\
\hline
CUHK DeepID-Net & 2014 & 2nd & $40.7\%$ & ImageNet $1$k & ? & CNN \\
\hline
GoogLeNet & 2014 & 1st & $43.9\%$ & ImageNet $1$k & 6 & CNN\\
\hline
\end{tabular}
\caption{Detection performance}
\label{detfinal}
\end{table}

We first report the top detection results and show the progress since the first edition of the detection task. Compared to the 2013 result, the accuracy has almost doubled. The top performing teams all use Convolutional Networks. 
We report the official scores in Table \ref{detfinal} and common strategies for each team: the use of external data, ensemble models or contextual models. The external data is typically the ILSVRC12 classification data for pre-training a model that is later refined on the detection data. Some teams also mention the use of the localization data. Since a good portion of the localization task bounding boxes are not included in the detection dataset, one can pre-train a general bounding box regressor with this data the same way classification is used for pre-training. The GoogLeNet entry did not use the localization data for pretraining.

\begin{table}
\centering
\begin{tabular}[H]{|l||l|c|c|}
\hline
{\bf Team} & {\bf\ \ mAP\ \ } & {\bf Contextual model} & {\bf Bounding box regression} \\
\hline
Trimps-Soushen & $31.6\%$ & no & ? \\
\hline
Berkeley Vision &  $34.5\%$ & no & yes \\
\hline
UvA-Euvision &  $35.4\%$ & ? & ? \\
\hline
CUHK DeepID-Net2 & $37.7\%$ & no & ? \\
\hline
GoogLeNet & $38.02\%$ & no & no \\
\hline
Deep Insight & $40.2\%$ & yes & yes \\
\hline
\end{tabular}
\caption{Single model performance for detection}
\label{detsinglemodel}
\end{table}

In Table \ref{detsinglemodel}, we compare results using a single model only. The top performing model is by Deep Insight and surprisingly only improves by 0.3 points with an ensemble of 3 models while the GoogLeNet obtains significantly stronger results with the ensemble.

\section{Conclusions}
Our results seem to yield a solid evidence that approximating the expected optimal sparse structure by readily available dense building blocks is a viable method for improving neural networks for computer vision. The main advantage of this method is a significant quality gain at a modest increase of computational requirements compared to shallower and less wide networks. Also note that our detection work was competitive despite of neither utilizing context nor performing bounding box regression and this fact provides further evidence of the strength of the Inception architecture.
Although it is expected that similar quality of result can be achieved by much more expensive networks of similar depth and width, our approach yields solid evidence that moving to sparser architectures is feasible and useful idea in general. This suggest promising future work towards creating sparser and more refined structures in automated ways on the basis of~\cite{arora2013bounds}.

\section{Acknowledgements}
We would like to thank Sanjeev Arora and Aditya Bhaskara for fruitful discussions on~\cite{arora2013bounds}. Also we are indebted to the DistBelief~\cite{dean2012large} team for their support especially to Rajat Monga, Jon Shlens, Alex Krizhevsky, Jeff Dean, Ilya Sutskever and Andrea Frome. We would also like to thank to Tom Duerig and Ning Ye for their help on photometric distortions. Also our work would not have been possible without the support of Chuck Rosenberg and Hartwig Adam.

\bibliographystyle{plain}
\bibliography{bibliography}

\end{document}